\documentclass[runningheads]{llncs}

\usepackage[review,year=2024,ID=283]{accv}

\usepackage{accvabbrv}

\usepackage{graphicx}
\usepackage{booktabs}
\usepackage{soul}
\usepackage{wrapfig}
\usepackage{makecell}
\usepackage{bm}
\setstcolor{red}

\usepackage[accsupp]{axessibility}  

\usepackage[pagebackref,breaklinks,colorlinks,citecolor=accvblue]{hyperref}

\usepackage{orcidlink}

\usepackage{fancyhdr}
\begin{document}

\title{Online Pseudo-Label Unified Object Detection for Multiple Datasets Training} 

\titlerunning{Abbreviated paper title}

\author{XiaoJun Tang\inst{1} \and
Jingru Wang\inst{1} \and
Zeyu Shangguan\inst{1} \and Darun Tang\inst{1} \and Yuyu Liu\inst{1}}

\authorrunning{F.~Author et al.}

\institute{BOE Technology Group Co., Ltd, Beijing, China}

\maketitle

\begin{abstract}
  The Unified Object Detection (UOD) task aims to achieve object detection of all merged categories through training on multiple datasets, and is of great significance in comprehensive object detection scenarios.
  In this paper, we conduct a thorough analysis of the cross datasets missing annotations issue, and propose an \textbf{Online Pseudo-Label Unified Object Detection} scheme.
  Our method uses a periodically updated teacher model to generate pseudo-labels for the unlabelled objects in each sub-dataset.
  This periodical update strategy could better ensure that the accuracy of the teacher model reaches the local maxima and maximized the quality of pseudo-labels. In addition, we survey the influence of overlapped region proposals on the accuracy of box regression. We propose a category specific box regression and a pseudo-label RPN head to improve the recall rate of the Region Proposal Network (PRN).
  Our experimental results on common used benchmarks (\eg COCO, Object365 and OpenImages) indicates that our online pseudo-label UOD method achieves higher accuracy than existing SOTA methods.
  \keywords{Unified Object Detection \and Semi-supervised learning \and Pseudo Label}
\end{abstract}

\section{Introduction}
\label{sec:intro}

\begin{figure}[t]
\begin{subfigure}{0.45\linewidth}
    \centering
    \includegraphics[width=0.9\linewidth]{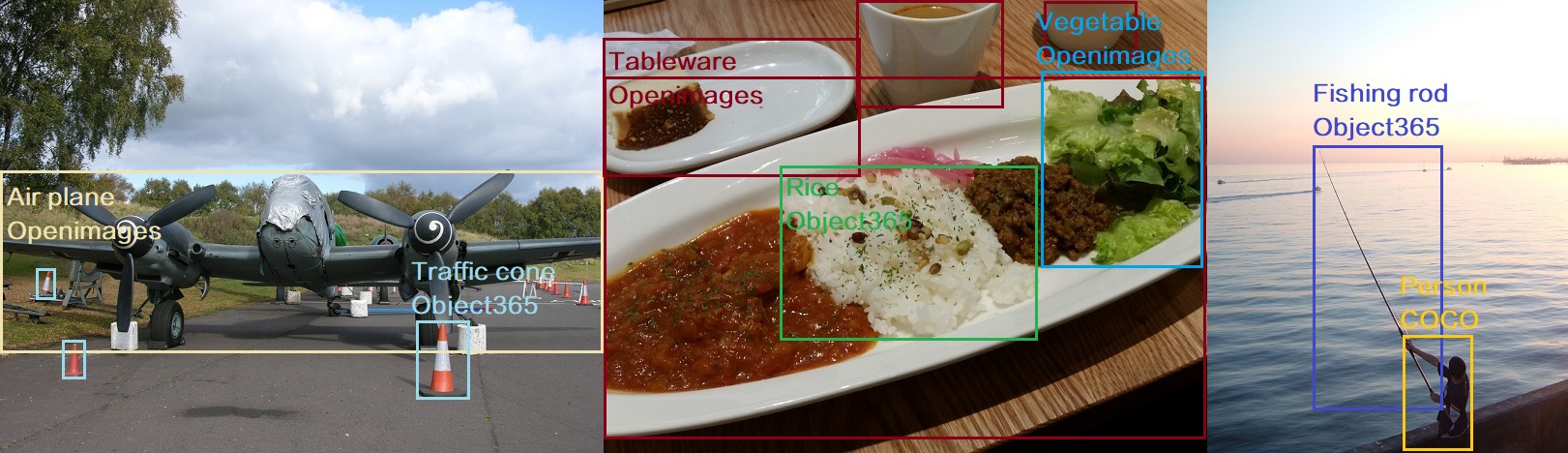}
    \caption{Different label spaces of different datasets}
    \label{fig:1_a}
\end{subfigure} 
\begin{subfigure}{0.55\linewidth}
    \centering
    \includegraphics[width=0.9\linewidth]{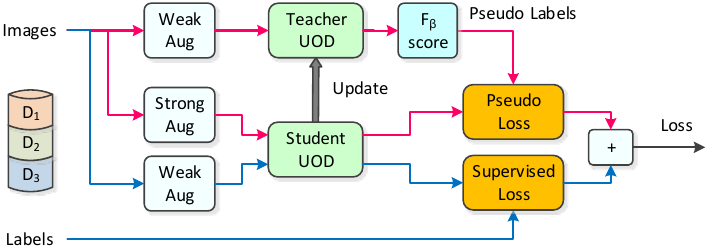}
    \caption{Online Pseudo-Label Unified Object Detection}
    \label{fig:1_b}
\end{subfigure}
\caption{(a) Ambiguous background problem: some categories in one dataset are not annotated in other datasets. (b) Our proposed Online Pseudo-Label UOD (OPL-UOD) uses a periodically updated teacher model to generate pseudo-labels of the unlabelled objects in the datasets, which enables the model training to obtain more cross datasets annotations and improved the UOD performance. Blue arrows indicate the manual label supervision training process, and red arrows indicate the pseudo-label supervision training process.}
\label{fig:1}
    \vspace{-10pt}
\end{figure}

Object detection task aims to detect various categories of objects in the wild. The training process requires corresponding scenarios images annotated with bounding boxes. Existing public datasets typically only annotate limited categories of objects (\eg COCO \cite{ref04}, WIDER FACE \cite{ref14}, SCUT \cite{ref07}, Object365 \cite{ref09}, OpenImages \cite{ref03}).
However, when need to build a single large-scale (\ie unified) dataset by fusing these public datasets, it requires vast extra annotation work. For example, if dataset $A$ consists of category $a,b$, and dataset $B$ contains category $b$, when we fuse $A$ and $B$ to a unified dataset $C = A+B$ for training, ideally all objects in $A$ of category $c$ should be fully labeled, but this process demand huge human labor costs.
Therefore, various advanced methods \cite{ref10,ref11,ref13,ref16,ref20,Shinya2021USBUO,Zhang2023Uni3DAU,Wu2023TowardsL3,Chen23ScaleDet} have been proposed to achieve an effective training on all categories through a strategic combination of multiple datasets, and avoid the annotating burden, also known as \textbf{unified object detection (UOD)}~\cite{Chen2023ScaleDetAS,Lin2022UniversalOD,Ye2023CascadeDETRDI,Cai2022BigDetectionAL,Wang2023DetectingEI}.

The two major challenges in the UOD task are: \textbf{taxonomy difference} \cite{ref16, ref20} and \textbf{background ambiguity} \cite{ref10, ref11, ref20}. The taxonomy difference issue means the name of certain categories among different dataset are various, \eg the `person' category in a dataset might be named as `pedestrian' in another one; the `football' category in a dataset means American football, but means soccer in another one. The general resolutions include: manually merging the categories of multiple datasets to form a unified label space \cite{ref10, ref11}, or train the model to learn a unified label space \cite{ref16}, or adaptly encoding the category names with word embedding \cite{ref20}. As for the background ambiguity issue, given a unified label space, each dataset only has a subset of the overall categories fully annotated, and the remaining categories are omitted, for example, in Fig.~\ref{fig:1_a}, some categories in one dataset are not annotated in other datasets. When using general object detection methods to implement naive multiple dataset combination training in this case, the unlabelled categories are tended to be mistakenly identified as background during training, which significantly reduced the accuracy of the model. To alleviate this problem, current UOD methods apply multiple binary sigmoid operations instead of a softmax function to predict the class scores, so that the loss of each class prediction could be calculated separately on different datasets\cite{ref16}. In addition, semi-supervised methods are getting popular\cite{ref10, ref11}. They use a pre-trained teacher models to generate offline pseudo-labels, and consequently require a two-step training scheme. In this paper, we concentrate on the background ambiguity issue and propose an online pseudo-label UOD scheme with periodically updated teacher models, which only required \textbf{single-step} training. As seen in Fig.~\ref{fig:1_b},  during the training process, the proposed periodical teacher update strategy ensured that the accuracy of the teacher model reached local maxima, and both the teacher model and the student model could mutually promote each other.

\begin{figure}[t]
    \centering
    \includegraphics[width=0.7\linewidth]{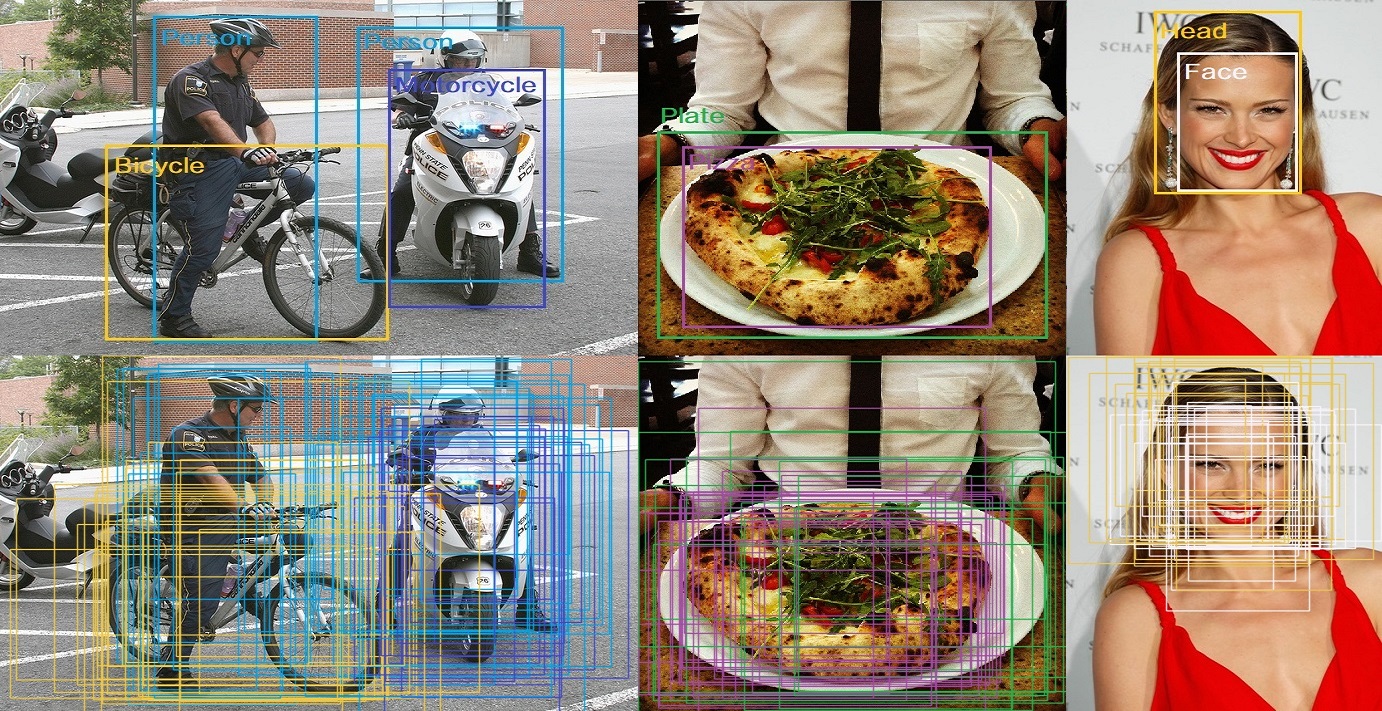}
    \caption{The overlapped boxes problem. The top row demonstrates that the annotation boxes of certain classes are prone to overlap. The bottom row shows the positive region proposals, which are marked with the corresponding colors to represent different ground truth categories.
    }
    \label{fig:2}
\end{figure}

In addition, we notice that as the number of the annotation categories increases, the overlapped boxes problem becomes non-negligible, which significantly degrade the accuracy of the box regression. 
We use the classic two-stage object detector to visualize the outputs of the Region Proposal Network (RPN), as the bottom row of Fig.~\ref{fig:2} demonstrated, there are a large number of region proposals associated with different categories overlap with each other, such as heads and faces, motorcycles and people, plates and pizza \etc. Considering that the UODs use multiple sigmoid binary class predictors, a single region proposal may produce multiple predictions of different categories. Therefore, it is necessary to design a category specific box regressor in the RCN of UOD. The category specific box regression could be implemented easily in early detectors, such as Faster RCNN \cite{ref08} and SSD \cite{ref05}, however, it conflicts with the more accurate CascadeRCNN detector \cite{ref02}, which has a multi-stage RCN architecture, which means the latter RCN stage needs to use the class unspecific box regression of the previous RCN stage for feature resampling. Based on this, we design a CascadeRCNN compatible category specific box regression structure to improve the box regression accuracy.

Given the situation that the RPN of the classic UOD network only conduct foreground-background classification, the unlabeled foreground objects would be treated as background in a dataset. However, other datasets may have these categories well-annotated; this phenomenon is also called \textbf{cross-datasets missing annotations problem}. To alleviate this problem, we designed the pseudo-label RPN training scheme, details will be discussed in the following sections.

Our main contributions includes:
\begin{itemize}
    \item To generate more accurate pseudo-labels, we propose a novel \textbf{online pseudo-label UOD} training scheme with periodically updated teacher models.
    \item To alleviate the overlapped boxes problem, we propose the \textbf{category specific box regression}, which obviously improved the box regression accuracy of UOD.
    \item For the background ambiguity issue, we propose the \textbf{pseudo-label RPN training}, which significantly improve the recall rate of the RPN head.
    \item To the best of our knowledge, this is the first online pseudo-label UOD method, and can outperform the offline pseudo-label UOD.
    \item Our method outperforms SOTA UOD detectors \cite{ref16, ref20} on the COCO, Object365 and OpenImages datasets.
\end{itemize}

\section{Related Work}
\textbf{Multiple Datasets Detection.} Multiple datasets training is an effective method to improve model robustness. It has been applied in semantic segmentation \cite{ref23, ref24}, depth estimation \cite{ref25}, and stereo matching \cite{ref26} \etal As far as multiple datasets object detection is considered, both different semantic concepts of categories and definitions of objects and background among different datasets need to be unified. Wang \etal \cite{ref13} design a partitioned detector with multiple RCN heads. Each head is actually trained on the corresponding dataset specifically. During model evaluation, the model needs to know which dataset the test image came from, thus made the detector unable to be used in actual application scenarios. Zhao \etal \cite{ref10} generate offline pseudo-labels to alleviate the cross dataset missing annotations problem and improved the mAP score significantly. Xu \etal \cite{ref17} propose a transferable graph R-CNN to model the class relations and improved the accuracy of the partitioned detector. But this partitioned detector again would produce duplicated outputs for the same object appearing in different datasets. Zhou \etal \cite{ref16} propose a simple and effective UOD architecture, and designed an automatic method to unify label spaces of multiple datasets. Their UOD architecture achieves SOTA mAP scores on large datasets. However, the training process neglect the utilization of pseudo-labels. Meng \etal \cite{ref20} propose the Detection Hub, which semantically aligned the categories across datasets by replacing one-hot category representations with word embedding, and achieved SOTA performance on wide variety of datasets. Similarly, the Detection Hub does not use pseudo labeling techniques. This paper propose a novel UOD scheme with the online pseudo-label, the category specific box regression for CascadeRCNN and the pseudo-label RPN training, which improve the UOD accuracy effectively.

\noindent \textbf{Semi-Supervised Learning.} The existing pseudo-label methods of the UOD training are all offline. While online pseudo-label methods have been widely used in the semi-supervised learning (SSL). In SSL, models are trained from a small amount of labeled data and a large amount of unlabeled data. As one of the most successful SSL algorithms, the online pseudo-label method uses teacher models to automatically annotate unlabeled data for training student models, and gained higher accuracy. The pseudo-labeling has been applied in image classification tasks. \cite{ref01, ref15} generate annotations on weakly augmented data and then apply strong augmentation on the training data with pseudo-labels. They aim to regularize model to be robust to small perturbation on model inputs or hidden states. To improve the quality of pseudo-labels, Tarvainen \etal \cite{ref12} proposed that the teacher model should be updated by an EMA (Exponential Moving Average) \cite{ref22} method instead of barely replicating the student model. According to these works, Liu \etal \cite{ref06} adopt SSL for the object detection. They subtly use focal loss to address the issue of unbalanced pseudo-labels. Since semi-supervised object detection requires to filter false-positive predicted bounding boxes using confidence scores, they use an empirical value as threshold. However, while their teacher model is gradually promoted, this threshold might become inappropriate during training. Furthermore, common datasets use in UODs typically presented a long tail distribution, it is reasonable to set specific thresholds for classes with different amount of training data. Tanaka \etal \cite{ref11} propose to determine the threshold for each class by maximizing score of both ground truth and pseudo-labels with a fixed teacher model. Wang \etal \cite{ref18} propose a GMM method to determine the threshold for each class during training and obtained higher accuracy than using fixed thresholds. Different from the above SSL methods, our proposed online pseudo-label UOD periodically updated the teacher model to ensure its accuracy reach local maxima, so that higher-quality pseudo-labels could be obtained during UOD training.

\section{The Proposed Approach}
\subsection{Preliminary of the Baseline UOD}
Given $N$ separate datasets $D = \{D_{0}$, $D_{1}$, ... $D_{N-1}\}$, with corresponding label spaces $L_{0}$, $L_{1}$, $L_{N-1}$, following \cite{ref16}, we merge them into a unified label space $L= L_{0}\bigcup L_{1}\bigcup ...\bigcup L_{N-1}$. Each label space is a subset of the unified label space $L_{i}\subseteq L$, and different label spaces are allowed to have common categories $L_{i}\bigcap L_{j}\ne \phi $. The goal is to train an UOD model on $D$ with label space $L$.

\begin{figure}[t]
    \centering
    \includegraphics[width=0.6\linewidth]{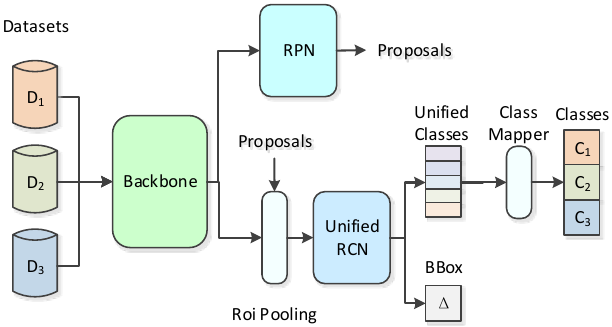}
    \caption{The baseline UOD structure.}
    \label{fig:3}
\end{figure}

Fig.~\ref{fig:3} illustrates the structure of the UOD baseline~\cite{ref16}, which uses a two-stage object detector with a shared backbone, a RPN head and a unified RCN head. The last Fully Connected (FC) layer of the RCN head calculates all class predictions among the unified label space $L$. During training, the input images are sampled from the unified datasets, while each batch can only have images sampled from a specific dataset $D_{i}$. The category predictions of $L$ are mapped to the label space $L_{i}$ of the corresponding dataset $D_{i}$, so that the UOD model could be trained on different datasets separately. The classification loss of the baseline UOD is shown in Eq.~\ref{formula01}, where $I$ is an image sampled from the dataset $D_{i}$, and $B$ denotes the box annotations of $I$. The region proposals are generated by the RPN head and are represented by $R\left ( I,B \right )$. For each region proposal $r$, the RCN head predicts all category scores $p_{c}^{L}\left ( r,\theta  \right )$ of the unified label space $L$, from which the class scores $p_{c}^{L_{i}}\left ( r,\theta  \right )$ of the sub-space $L_{i}$ are selected. We use $\theta$ to represent the UOD model parameters. Therefore, the Binary Cross Entropy (BCE) loss could be calculated with $p_{c}^{L_{i}}\left ( r,\theta  \right )$ and the ground truth class label $q_{c}\left ( r \right )$. Since classification scores outside of $p_{c}^{L_{i}}\left ( r,\theta  \right )$ are not used in the loss calculation; the cross dataset missing annotation problem do not affect the classification loss. In this way, the background ambiguity is effectively avoided in the RCN head training.

\begin{equation}
{
\label{formula01}
    L_{c}= \sum_{r\sim R\left ( I,B \right ) }BCE\left [ p_{c}^{L_{i}}\left ( r,\theta  \right )  ,q_{c}\left ( r \right ) \right ]
    }
\end{equation}

\begin{equation}
{
\label{formula02}
    L_{c}^{p^{+}}= \sum_{r \sim R\left ( I,B_{h}^{ps} \right ) } BCE \left [ p_{c}^{\tilde{L_{i}}} \left ( r, \theta   \right ) ,q_{c}^{p} \left ( r \right )  \right ]  \cdot  1\left ( q_{c}^{p}\left ( r \right )  \ge 0 \right ) 
    }
\end{equation}

\begin{equation}
{
\label{formula03}
    L_{c}^{p^{-}}= \sum_{r \sim R\left ( I,B_{l}^{ps} \right ) } BCE \left [ p_{c}^{\tilde{L_{i}}} \left ( r, \theta   \right ) ,q_{c}^{p} \left ( r \right )  \right ]  \cdot  1\left ( q_{c}^{p}\left ( r \right ) < 0 \right ) 
    }
\end{equation}

\subsection{Online Pseudo-Label Scheme}
Although existing offline pseudo-label training methods are effective in improving the accuracy of the UOD task~\cite{ref10, ref11}, it requires pre-training a teacher model for pseudo-label generation. This two-step training operation consumes more training time. In addition, during offline pseudo-label training, the teacher model is fixed and thus could not be updated with the student model, which hinders the improvement of the pseudo-labels. In this section, we are committed to research online pseudo-label methods for the UOD training, which only requires \textbf{one stage} of training.

\noindent \textbf{Pseudo-label classification loss.} Following \cite{ref10, ref11}, we also use a high threshold $T_{h}$ and a low threshold $T_{l}$ to select pseudo-labels generated by the teacher model. The pseudo-labels with detection scores higher than $T_{h}$ are treated as positive objects. On the contrary, the pseudo-labels with detection scores lower than $T_{l}$ are treated as negative background. Otherwise, the proposals are ignored. A positive pseudo-label classification loss $L_{c}^{p+}$ could be calculated following Eq.~\ref{formula02}, in which $B_{h}^{ps}$ is the pseudo-label generated by $T_{h}$, and the region proposals are represented by $R\left ( I,B_{h}^{ps} \right )$, $p_{c}^{\tilde{L_{i}} }\left ( r,\theta \right )$ represents the class scores outside of $L_{i}$, $q_{c}^{p}\left ( r \right )$ is the pseudo-label class label. $1\left ( q_{c}^{P}\left ( r \right ) \ge 0 \right )$ select the positive region proposals for loss calculation. Similarly, a negative pseudo-label classification loss $L_{c}^{p-}$ could be calculated through Eq.~\ref{formula03}, in which $B_{l}^{ps}$ is the pseudo-label generated by $T_{l}$, and $1\left ( q_{c}^{P}\left ( r \right ) < 0 \right )$ selected the negative region proposals for loss calculation. The overall classification loss is calculated as $L_{c}$ + $L_{c}^{p+}$ + $L_{c}^{p-}$.

\noindent \textbf{Analysis of classic EMA teacher updating.} For online pseudo-label UOD model training, the teacher model would be first initialized with weights of the student model after a certain period of training, and needs to be continuously updated once better student model is available.

To update the teacher model, we first try to adopt the standard EMA \cite{ref22} teacher model updating of SSL methods. The EMA updating is illustrated as Eq.~\ref{formula04}, in which $\alpha$ is the decay parameter. Experimental results indicate that the mAP score of the online pseudo-label UOD with the EMA update is much lower than that of the offline pseudo-label training. An important reason is that the offline pseudo-label training have a fixed pre-trained teacher model, and pseudo labels are obtained throughout the entire student model training process. While the online pseudo-label training only obtain the teacher model in the later stage of training.

\begin{equation}
{
\label{formula04}
    W_{n}^{Teacher}=  \alpha \cdot W_{n-1}^{Teacher} +  \left ( 1- \alpha  \right ) W_{n}^{Student}
    }
\end{equation}

\begin{figure}[t]
    \centering
    \includegraphics[width=0.65\linewidth]{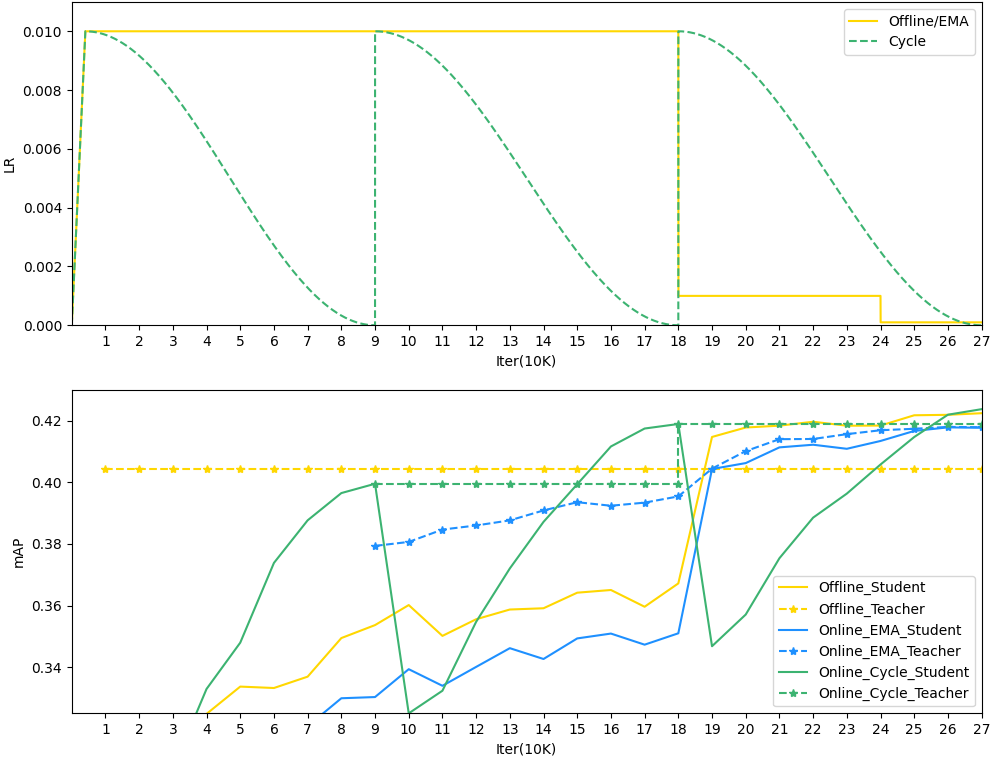}
    \caption{The top figure indicates that both the offline pseudo-label training and the EMA online pseudo-label UOD uses a step learning rate schedule (yellow curve), and the proposed online pseudo-label UOD used a cosine learning rate schedule (green curve). The bottom figure shows the mAP scores on the COCO-Split5 datasets. After adopting the periodical teacher updating, the mAP score of the cycle student model reaches the local maxima at minimum points of the cosine learning rate. The mAP score of the cycle teacher model is much higher than the EMA teacher.}
    \label{fig:4}
    \vspace{-10pt}
\end{figure}

\begin{figure}[t]
    \centering
    \includegraphics[width=0.6\linewidth]{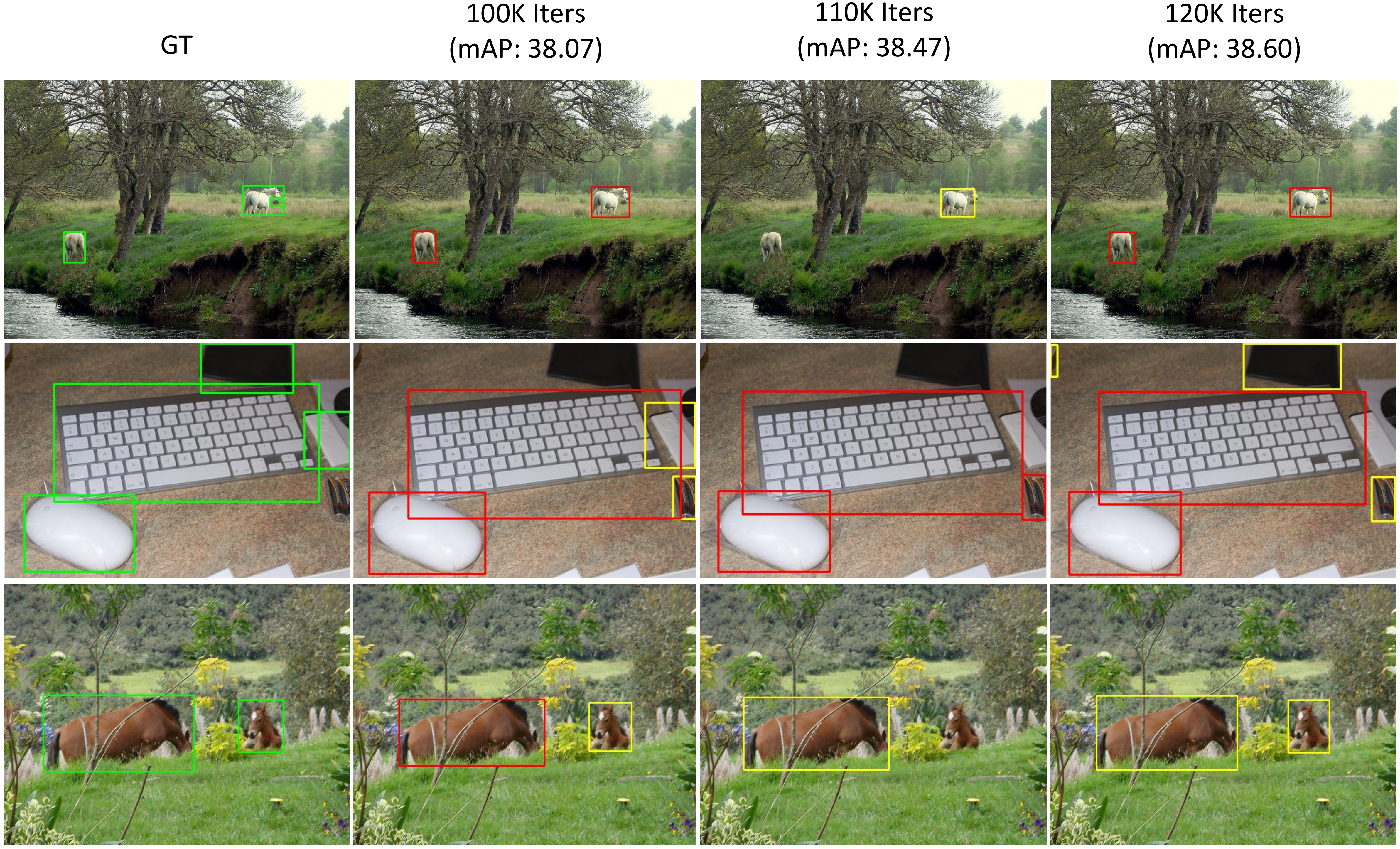}
    \caption{Pseudo-labels of different EMA teacher models on the COCO, Object365 and OpenImages datasets.}
    \label{fig:5}
\end{figure}

 Besides, we found another two problems of EMA updating, which are able to be improved. (i) The EMA offline pseudo-label training has the learning rate dilemma. On the one hand, a large learning rate is needed: considering that the EMA operation achieves the effect of improving model accuracy by averaging different input models, a low learning rate would reduce the difference between models with adjacent training time, thereby reducing the effectiveness of the EMA in improving the model accuracy, it also promotes the convergence of model. On the other hand, a small learning rate is needed for the accuracy of the student model to reach local maxima, which helps obtain a better teacher model. As shown by the yellow curve in top of Fig.~\ref{fig:4}, both the offline pseudo-label training and the EMA online pseudo-label UOD uses the step learning rate schedule. And the bottom of Fig.~\ref{fig:4} demonstrates the mAP scores on the COCO-Split 5 datasets (we will introduce this datasets in Section 4). For the offline pseudo-label training, the pre-trained teacher model is fixed and has a constant mAP score. Before the 180K iteration, due to the high learning rate, the mAP scores of the EMA student model is low. The mAP score of the EMA teacher model is lower than the offline teacher model obviously, which lead to worse pseudo-labels. At the 180K iteration, as the learning rate decayed, the mAP scores of the EMA student model gain remarkable improvement. This is because reducing the learning rate could make the model accuracy reach a local maximum, which is common in the step learning rate schedule training. After 180K iterations, the mAP score of the EMA teacher exceeds the offline teacher model. During this period, because of the small learning rate, the model training optimization slow down. Besides, the effectiveness of EMA is weakened as the gap of mAP between the EMA teacher model and EMA student model has been narrowed. The final mAP score of the EMA student model is lower than that of the offline student model. (ii) The teacher model for EMA online pseudo-label UOD is unstable. Fig.~\ref{fig:5} compares the pseudo-labels annotated by different EMA teacher models on the COCO, Object365 and OpenImages datasets. In this figure, the first column shows the GT annotations, and the following columns shows the pseudo-labels annotated by teacher models of the 100K, 110K and 120K iteration. The green boxes represent ground truth boxes, red boxes represent pseudo-labels with scores higher than the high threshold $T_{h}$ and yellow boxes represent those with scores higher than the low threshold $T_{l}$. The EMA updating improves the mAP score of the teacher model continuously, but for individual images, the accuracy of pseudo-labels fluctuated. The pseudo-labels that could be detected by the 100K teacher model might be missed by the latter 110K teacher model.

\noindent \textbf{Proposed periodical teacher updating.} To alleviate the above two problems, we propose the periodical teacher model updating for the online pseudo-label UOD. As the top part of Fig.~\ref{fig:4} illustrates, we apply the cosine learning rate schedule (green curve) for the online pseudo-label UOD. For each training cycle, we start with a large learning rate to accelerate the convergence of the student model, then gradually decay the learning rate to the lowest point to improve the accuracy of the student model quickly, and the teacher model is updated at the end of each cycle. The benefits are: (i) the teacher model is updated at minimum points of the learning rate, which enables its accuracy reach the local maxima, so as to improve the quality of pseudo-labels; (ii) due to the fact that the model is updated at only a few time points, it is convenient to use the computationally intensive $F_{\beta}$ score method \cite{ref11} to calculate the optimal threshold for different categories specifically.

\subsection{Category specific Box Regression for CascadeRCNN}

\begin{figure}[t]
    \centering
    \includegraphics[width=0.8\linewidth]{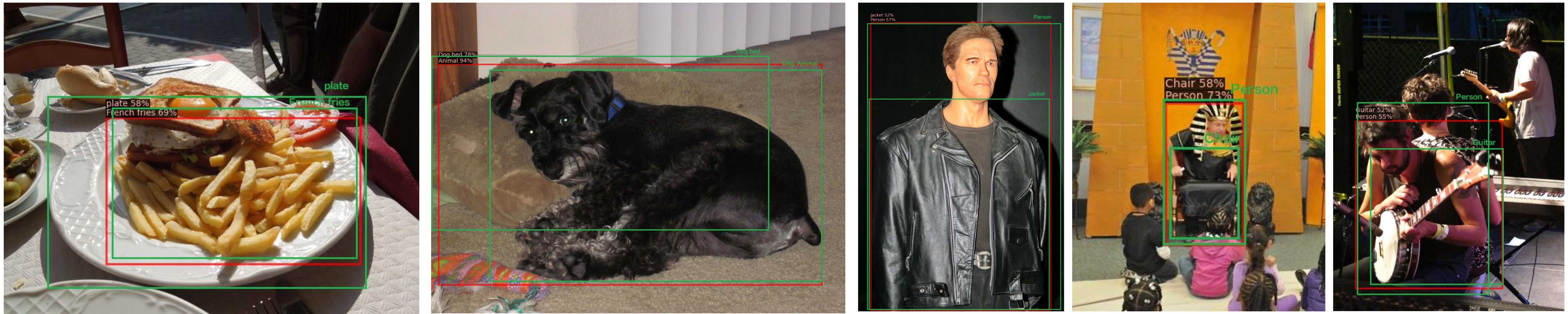}
    \caption{Examples of one proposal multiple class outputs with the score threshold 0.5. The proposal near the overlapped objects have two class scores larger than 0.5. The two objects share a box regression in the baseline UOD, and is drawn as a red box. The corresponding ground truth boxes are drawn as green boxes, which are obviously different from the box regression outputs.}
    \label{fig:6}
\end{figure}

\begin{figure}[t]
    \centering
    \includegraphics[width=0.5\linewidth]{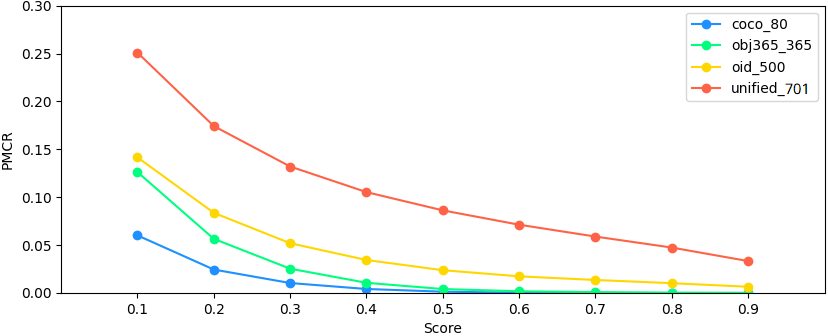}
    \caption{Proposal Multiple Class Ratio (PMCR) on the COCO (80 Classes), Object365 (365 Classes), OpenImages (500 Classes) and the unified datasets (701 Classes). The larger the number of dataset categories, the higher the PMCR.}
    \label{fig:7}
\end{figure}

\noindent \textbf{Analysis of overlapped boxes problem.} As stated in ~\ref{sec:intro}, box predictor for different categories shares one box regression in the existing UOD models, which results in locating errors. As illustrated in Fig.~\ref{fig:6}, for quantitative analysis, we use the baseline UOD \cite{ref16} on the COCO, Object365 and OpenImages datasets, and test on the validation datasets. For each test image $I$, the RPN head extracted 1,000 proposals and sent them to the RCN head to calculate the class scores of different categories. For a given score threshold $t$, the number of proposals with at least one class score greater than $t$ is denoted as $P_{1}\left ( I, t \right ) $, and the number of proposals with two or more class scores greater than $t$ is denoted as $P_{2}\left ( I, t \right ) $. Thus the Proposal Multiple Class Ratio could be calculated following Eq.~\ref{formula05} , which has a value range of $0 \sim 1$. Fig.~\ref{fig:7} illustrates that the value of PMCR would increase as the number of categories raised.

\begin{equation}
{
\label{formula05}
    PMCR\left ( t \right )  = \frac{\sum_{I}P_{2}\left ( I,t \right )  }{\sum_{I}P_{1}\left ( I,t \right )} 
    }
\end{equation}

\begin{figure}[t]
    \centering
    \includegraphics[width=0.5\linewidth]{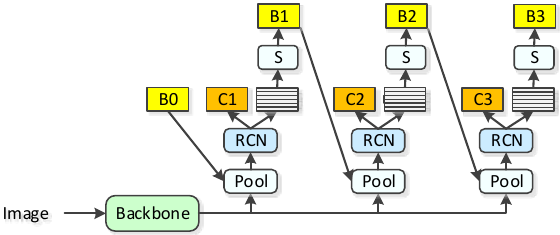}
    \caption{Category specific box regression for CascadeRCNN.}
    \label{fig:8}
\end{figure}

\noindent \textbf{Category specific box regression for CascadeRCNN.} The category number of the unified label space is usually very large in UOD tasks, and the overlapped boxes problem became non-negligible. Therefore, we propose the category specific box regression based on CascadeRCNN in this section. It allows the model to output category specific box regressions for difference categories. The standard RCN head uses the category unspecific box regression of previous stages to resample the features of the region proposals in the next stage. In Fig.~\ref{fig:8}, the proposed category specific box regression RCN head outputs category specific boxes. We select the box with the highest class prediction score in both the training and inference process.

\subsection{Pseudo-label RPN training}
The RPN head is responsible for generating candidate region proposals, indicating whether the corresponding anchor box has an object or background. As different datasets have different background definitions \cite{ref10}, the RPN training also suffer from the ambiguous background problem. Therefore, we also generate the pseudo-labels for this stage.

\section{Experiments}
\textbf{Datasets.} We evaluate the performance of our proposed OPL-UOD on COCO split 5 datasets. We randomly divide COCO into $N\left ( N=5 \right )$ sub-datasets. At first, we split 80 categories into $N+1$ disjoint sub-category spaces with the category numbers of 14, 14, 13, 13, 13 and 13. The last sub-category space is merged in the other 5 sub-category spaces, and forms 5 sub-category spaces with the category numbers of 27, 27, 26, 26 and 26. The training images are also randomly split into 5 disjoint sub-training datasets. Each sub-training dataset and sub-validation dataset only retain the labels corresponding to the sub-category space.

For comparison with SOTA, we evaluate the OPL-UOD on three large datasets: COCO \cite{ref04}, Objects365 \cite{ref09} and OpenImages \cite{ref03}. Following the baseline UOD \cite{ref16}, we use the unified label space (701 categories) of multiple datasets generated by the automatic learning method.

In addition, we also test the category specific box regression on the WIDERFACE \cite{ref14} face detection dataset, and the SCUT \cite{ref07} head detection dataset. For all datasets except OpenImages, we use mAP at IOU thresholds 0.5 to 0.95 as evaluation metric. Following \cite{ref16}, for OpenImages, the official modified mAP@0.5 is used.

\noindent \textbf{Training details.} Based on the baseline UOD \cite{ref16} , we implement our models use the Cascade RCNN detector with ResNet50 backbone and FPN neck. In addition, we follow \cite{ref06} to replace the cross entropy loss with the Focal loss \cite{ref19} to alleviate the class imbalance.
For weakly data augmentation, we simply use random flip and scaling of the short edge in range [640, 800]. For strongly data augmentation, color jittering, grayscale, Gaussian blur and cutout patches are randomly added.
we use the SGD optimizer and set the base learning rate as 0.01. For the COCO, Object365 and OpenImages unified detection experiments, the models are trained with batch size 16 on 8 V100 GPUs. For other experiments, the models are trained with batch size 8 on 4 A10 GPUs. For the EMA updating, the decay parameter $\alpha$ is set as 0.9996. In order to reproduce the model training results, we use a fixed random seed in all model training.

\subsection{Category specific box regression}

\begin{figure}[t]
    \centering
    \includegraphics[width=80mm]{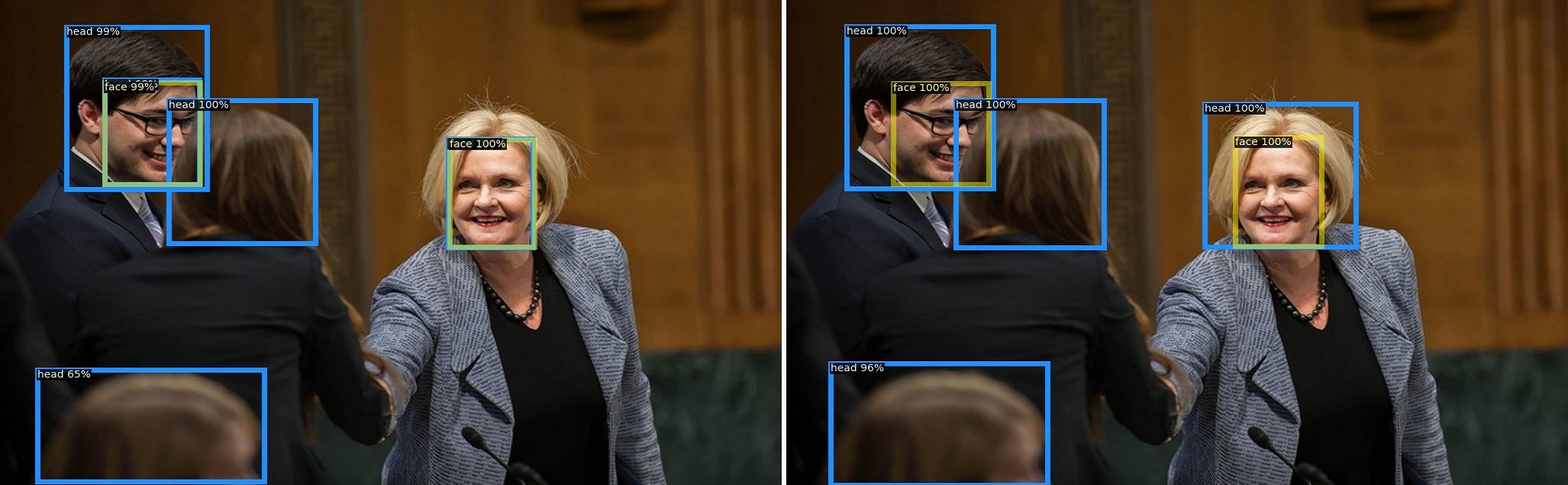}
    \caption{Comparison of different box regressions on the COCO, SCUT and WIDERFACE unified object detection. The left image is for the standard category unspecific box regression CascadeRCNN. The right image is for the proposed category specific box regression CascadeRCNN.}
    \label{fig:9}
\end{figure}

\begin{table}[t]
    \caption{Category specific box regression vs category unspecific box regression: mAP scores on the COCO, SCUT and WIDERFACE unified object detection.}
    \label{table_1}
    \centering
\scalebox{0.8}{
    \setlength{\tabcolsep}{2mm}
    \begin{tabular}{ccccc}
    \toprule
         Method & COCO & SCUT & WIDERFACE & mean \\
    \midrule
         Category unspecific & 41.8 & 47.5 & 32.8 & 40.7 \\
    \midrule
         Category specific & 
         $\textbf{42.5}_{\textcolor[RGB]{0,128,0}{(+0.7)}}$ &
         $\textbf{48.0}_{\textcolor[RGB]{0,128,0}{(+0.5)}}$ &
         $\textbf{33.3}_{\textcolor[RGB]{0,128,0}{(+0.5)}}$ &
         $\textbf{41.3}_{\textcolor[RGB]{0,128,0}{(+0.6)}}$ \\
    \bottomrule
    \end{tabular}
    }
\end{table}

We train the UOD models on COCO, SCUT \cite{ref07} and WIDER FACE \cite{ref14} datasets for 180K iterations to explore the effectiveness of the category specific box regression. Since faces and heads have a large number of overlapped boxes, the category unspecific box regression UOD inevitably would have a large positioning error. In Fig.~\ref{fig:9}, the left image shows detection boxes generated by the standard CascadeRCNN. The head box of the woman is incorrectly positioned on her face, and an additional head box is incorrectly positioned on the face of the man. The right image shows detection boxes generated by our category specific box regression CascadeRCNN, all head boxes and face boxes are accurately detected. Tab.~\ref{table_1} illustrates that the mean mAP score of the category specific box regression is 0.5 higher than that of the standard category unspecific box regression.

Category specific box regression is also effective in single dataset training. We train models on COCO, Objects365, and OpenImages respectively. Each dataset had a different number of classes. In addition we also trained a model on the unified dataset of the three datasets. All these models are trained for 180K iterations. Tab.~\ref{table_2} illustrated that the mAP scores of the category specific box regression on COCO, Objects365, OpenImages and the Unified dataset increased by 0.1, 0.3, 0.3 and 0.5 respectively. With increased number of classes, the boxes overlap more severely, and our method will achieve a greater improvement.

\begin{table}[t]
    \caption{The effectiveness of category specific box regression: mAP scores on different datasets.}
    \label{table_2}
    \centering
\scalebox{0.8}{
    \setlength{\tabcolsep}{2mm}
    \begin{tabular}{ccccc}
    \toprule
         Method & \makecell[c]{COCO \\ 80 Classes} & \makecell[c]{Object365 \\ 365 Classes} & \makecell[c]{OpenImages \\500 Classes} & \makecell[c]{Unified \\701 Classes}\\
    \midrule
         category unspecific & 42.0 & 22.7 & 63.0 & 39.6 \\
    \midrule
         Category specific & 
         $\textbf{42.1}_{\textcolor[RGB]{0,128,0}{(+0.1)}}$ &
         $\textbf{23.0}_{\textcolor[RGB]{0,128,0}{(+0.3)}}$ &
         $\textbf{63.3}_{\textcolor[RGB]{0,128,0}{(+0.3)}}$ &
         $\textbf{40.1}_{\textcolor[RGB]{0,128,0}{(+0.5)}}$ \\
    \bottomrule
    \end{tabular}
    }
\end{table}

\subsection{Pseudo-label RPN training}
We train the pseudo-label RPN models on the COCO split 5 datasets for 270K iterations. Tab.~\ref{table_3} indicates that RPN recall rates of the pseudo-label RPN training are obviously higher than the standard RPN training, which verified that the pseudo-label RPN training is able to improve the recall rate of the RPN head. However, in Tab.~\ref{table_4}, the pseudo-label RPN training had almost no improvement in mAP scores. This is because mAP scores of two-stage object detectors are mainly determined by the RCN head, and the influence of RPN head is moderate.

\begin{table}[t]
    \caption{RPN recall rates of pseudo-label RPN training on the COCO split 5 unified object detection.}
    \label{table_3}
    \centering
\scalebox{0.8}{
    \setlength{\tabcolsep}{0.8mm}
    \begin{tabular}{ccccccc}
    \toprule
         Method & Split1 & Split2 & Split3 & Split4 & Split5 & Mean \\
    \midrule
         Standard RPN & 50.3 & 46.9 & 48.8 & 50.8 & 52.5 & 49.9\\
    \midrule
         Pseudo-label RPN & 
         $\textbf{50.4}_{\textcolor[RGB]{0,128,0}{(+0.1)}}$ & 
         $\textbf{47.6}_{\textcolor[RGB]{0,128,0}{(+0.7)}}$ & 
         $\textbf{49.3}_{\textcolor[RGB]{0,128,0}{(+0.5)}}$ &
         $\textbf{51.5}_{\textcolor[RGB]{0,128,0}{(+0.7)}}$ &
         $\textbf{53.1}_{\textcolor[RGB]{0,128,0}{(+0.6)}}$ &
         $\textbf{50.4}_{\textcolor[RGB]{0,128,0}{(+0.5)}}$ \\
    \bottomrule
    \end{tabular}
    }
\end{table}

\begin{table}[t]
    \caption{The pseudo-label RPN training have almost no improvement in mAP scores.}
    \label{table_4}
    \centering
\scalebox{0.8}{
    \setlength{\tabcolsep}{0.8mm}
    \begin{tabular}{ccccccc}
    \toprule
         Method & Split1 & Split2 & Split3 & Split4 & Split5 & Mean \\
    \midrule
         Standard RPN & 40.4 & 36.8 & 40.8 & 42.2 & 42.8 & 40.6\\
    \midrule
         Pseudo-label RPN & 
         $40.2_{\textcolor[RGB]{128,0,0}{(-0.2)}}$ &
         $\textbf{36.9}_{\textcolor[RGB]{0,128,0}{(+0.1)}}$ &
         $\textbf{40.9}_{\textcolor[RGB]{0,128,0}{(+0.1)}}$ &
         $\textbf{42.3}_{\textcolor[RGB]{0,128,0}{(+0.1)}}$ &
         $\textbf{42.9}_{\textcolor[RGB]{0,128,0}{(+0.1)}}$ &
         $\textbf{40.6}_{\textcolor[RGB]{128,128,128}{(+0.0)}}$ \\
    \bottomrule
    \end{tabular}
    }
\end{table}

\subsection{Online Pseudo-Labeling UOD (OPL-UOD)}
In this section, we evaluate the proposed online pseudo-label UOD scheme and compared it with other UOD methods. Tab.~\ref{table_5} illustrated the mAP scores of different UOD models trained on the COCO split 5 datasets for 270K iterations. The top 3 rows did not use pseudo labels. The baseline UOD \cite{ref16} had the mean mAP score of 39.9. The mean mAP score of the model using Focal loss is 0.5 higher. The mean mAP score of the model using cosine LR is slightly higher (+0.1) than that of the standard step LR. The middle 4th row is the offline pseudo-label training model, which use the  score \cite{ref11} to calculate the optimal threshold of the pre-trained teacher model. Its mAP score is 1.8 higher than the previous no pseudo-label models. However, the offline pseudo-label training required two step model training, and spent the longest model training time. The bottom 4 rows are online pseudo-label UOD models. The mAP score of the standard online EMA UOD is 1.3 higher than that of no pseudo-lable models, but is obviously lower (-0.5) than the offline pseudo-label training model. Our proposed OPL-UOD (90K iterations each cycle) achieved similar mAP scores as the offline pseudo-label training model. The mean mAP score of the OPL-UOD with pseudo-label RPN is 0.1 higher than that of the OPL-UOD. The mean mAP score of the OPL-UOD with pseudo-label RPN and category specific box regression is similar to the OPL-UOD with pseudo-label RPN. This is because the COCO has relatively less categories and the effectiveness of the category specific box regression is minor.

\begin{table}[t]
    \caption{mAP score comparison on the COCO split 5 unified object detection. (Take average value of 3 experiments)}
    \label{table_5}
    \centering
\scalebox{0.8}{
    \setlength{\tabcolsep}{1.7mm}
    \begin{tabular}{cccccccc}
    \toprule
         Method & Split1 & Split2 & Split3 & Split4 & Split5 & Mean & Time \\
    \midrule
         Baseline UOD & 39.5 & 36.2 & 40.3 & 41.6 & 41.7 & 39.9 & 1.0 \\
    \midrule
         Baseline+Focal & 40.1 & 36.8 & 40.9 & 42.1 & 42.3 & 40.4 & 1.0 \\
    \midrule
         Baseline+Focal+CosineLR & 40.2 & 36.8 & 40.8 & 42.2 & 42.8 & 40.5 & 1.0 \\
    \midrule
    \midrule
         Offline UOD & 41.7 & 38.7 & 42.3 & 44.1 & 44.8 & 42.3 & 3.3 \\
    \midrule
    \midrule
         Online EMA UOD & 41.1 & 38.4 & 41.6 & 43.6 & 44.3 & 41.8 & 1.9 \\
    \midrule
         OPL-UOD & 41.5 & 38.6 & 42.3 & 44.1 & 44.9 & 42.3 & 1.9 \\
    \midrule
         OPL-UOD+PRPN & 41.7 & 38.8 & \textbf{42.4} & 44.2 & 44.7 & 42.4 & 2.2 \\
    \midrule
         OPL-UOD+PRPN+MBox & \textbf{41.7} & \textbf{38.8} & 42.3 & \textbf{44.2} & \textbf{44.9} & \textbf{42.4} & 2.2 \\
    \bottomrule
    \end{tabular}
    }
\end{table}

\begin{table}[t]
    \caption{Compared to SOTA: mAP scores on the COCO, Object365 and OpenImages unified object detection.}
    \label{table_6}
    \centering
\scalebox{0.8}{
    \setlength{\tabcolsep}{2mm}
    \begin{tabular}{ccccc}
    \toprule
         Method & COCO & Obj365 & OID & Mean \\
    \midrule
         UOD \cite{ref16} (R50, 720K) & 45.4	& 24.4 & 66.0 & 45.3 \\
    \midrule
         UOD \cite{ref16} (R50, 2160K) & 45.7 & 25.4 & 67.8 & 46.3 \\
    \midrule
         Detection Hub \cite{ref20} (R50) & 45.3 & 23.2 & - & - \\
    \midrule
         Offline UOD & 47.1 & 27.2 & 69.5 & 47.9 \\
    \midrule
         OPL-UOD & \textbf{47.2} & 27.1 & 69.5 & 47.9 \\
    \midrule
         OPL-UOD+Mbox & 47.1 & \textbf{27.4} & \textbf{70.1} & \textbf{48.2} \\
    \midrule
         OPL-UOD+Mbox+PRPN & 47.1 & 27.1 & 70.1 & 48.1 \\
    \bottomrule
    \end{tabular}
    }
\end{table}

\subsection{Comparison with SOTA}
Tab.~\ref{table_6} illustrates the mAP scores of different models on COCO, Objects365 and OpenImages datasets. For fair comparison, all models use the ResNet50 backbone. The original UOD \cite{ref16} is trained for 720k iterations and has the mean mAP score of 45.3. We train the UOD for 2160k iterations and the mean mAP score increased to 46.3. The following Offline UOD and OPL-UOD models are also trained for the same 2,160k iterations. The Detection Hub \cite{ref20} is trained on the COCO, Object365 and VG \cite{ref21} at 1x schedule. Its mAP score on OpenImages is not available. The mAP scores on COCO and Objects365 are slightly lower than UOD \cite{ref16}, but they might improve with more training iterations. The mean mAP score of the offline pseudo-label training model is 1.6 higher than the UOD \cite{ref16}. The proposed OPL-UOD achieved similar mAP scores as the offline pseudo-label training model. The OPL-UOD with category specific box regression achieves the highest mean mAP score on the three large datasets. The pseudo-label RPN does not improve mAP scores on these datasets.

\subsection{Comparison of training time}
The last column of Tab.~\ref{table_5} illustrates the relative training time of different training schemes. It takes 26 hours to train the baseline UOD on 4 A10 GPUs, which is used as the unit benchmark time for calculating the relative training time of other models. The Focal Loss and CosineLR don’t affect the training time of the baseline UOD. While the training time of the offline pseudo-label training scheme is 3.3 times that of the baseline UOD. The online EMA and OPL-UOD have the same training time. The pseudo-label RPN has added a small amount of training time, while the category specific box regression has little impact on training time.

\section{Conclusion}
We propose an online pseudo-label unified object detector, which use a periodically updated teacher model to generate pseudo-labels for the unlabelled objects to alleviate the background ambiguity problem, and use a category specific box regressor to alleviate the overlapped boxes problem. Experimental results verify that our proposed periodical updating strategy is superior to the traditional EMA updating strategy, and achieve higher mAP scores than the offline pseudo-label training. We hope our periodically updated teacher model method could also be applied to future semi-supervised learning works. Furthermore, we also propose the category specific box regression for CascadeRCNN and the pseudo-label RPN training, which could improve the model performance.


%
%
\bibliographystyle{splncs04}
\bibliography{main}
\end{document}